\documentclass[letterpaper, 10 pt, journal, twoside]{IEEEtran}
\IEEEoverridecommandlockouts
\usepackage[utf8]{inputenc}
\usepackage[T1]{fontenc}
\usepackage{graphicx}
\usepackage{longtable}
\usepackage{wrapfig}
\usepackage{rotating}
\usepackage[normalem]{ulem}
\usepackage{amsmath}
\usepackage{amssymb}
\usepackage{capt-of}
\usepackage{hyperref}
\usepackage{subfig}
\usepackage{tabularx}
\usepackage{xcolor}
\usepackage{cite}
\usepackage{pgfplots}
\usepackage{tikz}
\usepackage{academicons}

\pgfplotsset{compat=1.16}
\usepgfplotslibrary{external}
\def\BibTeX{{\rm B\kern-.05em{\sc i\kern-.025em b}\kern-.08em T\kern-.1667em\lower.7ex\hbox{E}\kern-.125emX}}

\newcommand{\orcid}[1]{\href{https://orcid.org/#1}{\includegraphics[height=\fontcharht\font`A]{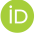}}}

\newcommand{\authors}{Daniel Mox\orcid{0000-0002-9513-4460}, \textit{Graduate Student Member, IEEE}, Vijay Kumar\orcid{0000-0002-3902-9391}, \textit{Fellow, IEEE},\\ and Alejandro Ribeiro\orcid{0000-0003-4230-9906}, \textit{Member, IEEE}}
\newcommand{\acknowledgments}{
  \thanks{Manuscript received September 9, 2021; accepted January 3, 2022. Date of publication January 27, 2022; date of current version March 22, 2022. This letter was recommended for publication by Associate Editor Guillaume Adrien Sartoretti and Editor M. Ani Hsieh upon evaluation of the reviewers’ comments. The work of Daniel Mox was supported by the National Science Foundation Graduate Research Fellowship under Grant DGE-1845298. This work was supported in part by ARL under Grant DCIST CRA W911NF-17-2-0181, in part by NSF under Grant CNS-1521617, in part by ARO under Grant W911NF-13-1-0350, in part by ONR under Grants N00014-20-1-2822 and N00014-20-S-B001, in part by Qualcomm Research, and in part by the NVIDIA Corporation through the donation of the DGX Station used for this research. (Corresponding author: Daniel Mox.)}
  \thanks{Daniel Mox and Vijay Kumar are with the General Robotics, Automation, Sensing and Perception (GRASP) Laboratory at the University of Pennsylvania, Philadelphia, PA 19104 USA (email: mox@seas.upenn.edu, kumar@seas.upenn.edu)}
  \thanks{Alejandro Ribeiro is with the Electrical and Systems Engineering Department at the University of Pennsylvania, Philadelphia, PA 19104 USA (email: aribeiro@seas.upenn.edu)}
  \thanks{This letter has supplementary downloadable material available at https://doi.org/10.1109/LRA.2022.3146524, provided by the authors.}
  \thanks{Digital Object Identifier 10.1109/LRA.2022.3146524}}
\author{\authors \acknowledgments}
\date{}
\title{Learning Connectivity-Maximizing Network Configurations}
\hypersetup{
 pdfauthor={\authors \acknowledgments},
 pdftitle={Learning Connectivity-Maximizing Network Configurations},
 pdfkeywords={},
 pdfsubject={},
 pdfcreator={Emacs 28.1 (Org mode 9.5.2)}, 
 pdflang={English}}
\begin{document}

\maketitle
\begin{abstract}
In this letter we propose a data-driven approach to optimizing the algebraic connectivity of a team of robots. While a considerable amount of research has been devoted to this problem, we lack a method that scales in a manner suitable for online applications for more than a handful of agents. To that end, we propose a supervised learning approach with a convolutional neural network (CNN) that learns to place communication agents from an expert that uses an optimization-based strategy. We demonstrate the performance of our CNN on canonical line and ring topologies, 105k randomly generated test cases, and larger teams not seen during training. We also show how our system can be applied to dynamic robot teams through a Unity-based simulation. After training, our system produces connected configurations over an order of magnitude faster than the optimization-based scheme for teams of 10-20 agents.
\end{abstract}
\section{Introduction}
\label{sec:org236c152}
\IEEEPARstart{E}{nsuring} a team of robots remains in contact with one another while accomplishing an objective has been the focus of a considerable amount of research in the robotics community. Indeed, most multi-robot systems operate on the underlying assumption that robots can freely exchange information in order to coordinate their actions. One common method for ensuring this condition is met considers the network graph induced by the spatial configuration of the robots coupled with an underlying communication model. This problem formulation emits a graph theoretic approach focused on maximizing or preserving the connectivity of the state dependent communication graph \cite{zavlanos2011graph,zavlanos2005controlling,zavlanos2007potential,stump2008connectivity,kim2005maximizing,de2006decentralized,ji2007distributed,zavlanos2008distributed,notarstefano2006maintaining,spanos2004robust,yang2010decentralized,ji2006distributed,dimarogonas2008decentralized} which can be solved in both a centralized \cite{zavlanos2011graph,zavlanos2005controlling,zavlanos2007potential,stump2008connectivity,kim2005maximizing} and decentralized manner \cite{zavlanos2011graph,de2006decentralized,ji2007distributed,zavlanos2008distributed,notarstefano2006maintaining,spanos2004robust,yang2010decentralized,ji2006distributed,dimarogonas2008decentralized}.

Communication graph connectivity is inherently a heuristic for network performance. This choice is based on the intuition that the more connected a network is the easier information flows through it. However, in practice, network performance depends on the underlying routing protocol and the extent to which maximizing graph connectivity directly translates to improved performance is unclear. More recent work has sought to bridge this gap by integrating aspects of wireless systems into the planning problem formulation. In \cite{yan2012robotic} the authors propose a mobile router solution that optimizes end-to-end bit error rate between a pair of nodes. Other work goes even further by combining the node positioning and packet routing problems \cite{mox2020mobile,stephan2017concurrent,fink2013robust,zavlanos2012network}. While these methods more accurately model the underlying wireless network they result in computationally expensive, coupled optimization problems that can be challenging to solve, even approximately.

In this work, we consider the problem of \emph{mobile wireless infrastructure on demand} introduced in \cite{mox2020mobile} wherein mobile relay nodes must be strategically positioned to facilitate communication between task-oriented robots whose actions may take them out of direct transmission range with one another. While many of the aforementioned approaches offer solutions to this problem, we lack a method that scales as the number of agents in the network grows. Indeed, centralized approaches to maximizing algebraic connectivity require solving costly optimization problems that become prohibitively slow for large teams. Decentralized variants rely on distributed optimization which is iterative in nature and requires significant time to converge. Likewise, the more sophisticated methods face similar scaling challenges just to find feasible solutions.

To address this problem, we propose a data-driven approach to maximizing connectivity that demonstrates attractive scaling characteristics compared with existing methods. In particular, we show how supervised learning coupled with convolutional neural networks (CNNs) can be used to learn how to provide mobile wireless infrastructure on demand, ensuring a multi-robot team forms a connected network. CNNs have been used in many robotics tasks such as perception, control, and planning \cite{bansal2020combining,akkaya2019solving,levine2016end,xiang2017posecnn,tamar2016value}. Similarly, images have emerged as a natural way to encode the spatial configuration of agents relative to each other and their environment \cite{tamar2016value,sartoretti2019primal,ndousse2021emergent}. A key insight motivating the use of learning in our case is that while training can be a long process, inference is inexpensive. In addition, the existing connectivity maximization solutions that aren't suitable for real-time applications can readily be used to generate expert training samples offline for supervised learning.

In this letter we present a novel approach to mobile infrastructure on demand utilizing supervised learning with CNNs. We show that our approach achieves comparable performance to the expert it learns from in a fraction of the time, generalizes to larger teams not seen during training, and operates effectively in dynamic teams of robots. The rest of this letter is organized as follows: Section \ref{sec:orgbedaad4} provides an overview of the problem and our approach, covering how we model communication between agents in \ref{sec:org92bc834}, how we generate training samples in \ref{sec:org0c6abe7}, and what CNN architecture we use in \ref{sec:orga21e708}, Section \ref{sec:org85ae6b9} contains results and analysis, and finally in Section \ref{sec:orga6e01a5} we provide concluding remarks.
\section{Methodology}
\label{sec:orgbedaad4}
Consider the scenario where a team of mobile robots seeks to accomplish a task requiring communication. Instead of creating and sustaining a wireless network in addition to completing their objective, these robots, referred to as task agents, assume the availability of communication infrastructure which is provided by a different set of robots, referred to as communication or relay agents. This team of relay agents positions itself in the environment and facilitates communication so that the task agents can go about their objective without considering the impact their actions have on their ability to exchange vital information. Our focus is on how the communication agents should be positioned.

More formally, given a set of \(N\) task agents with configuration denoted \(\boldsymbol{x}_T\in\mathbb{R}^{N\times 2}\), we seek to position a set of \(M\) communication agents \(\boldsymbol{x}_C\in\mathbb{R}^{M\times 2}\) so that the entire group \(\boldsymbol{x} = [\boldsymbol{x}_T; \boldsymbol{x}_C]\) maximizes the algebraic connectivity of the underlying communication graph. While more sophisticated placement algorithms exist \cite{mox2020mobile,stephan2017concurrent,fink2013robust}, they neither provide locally optimal solutions nor are they computationally suitable even for offline data generation. We target algebraic connectivity maximization since locally optimal configurations can be readily found using an optimization approach \cite{kim2005maximizing}. Unfortunately, such optimization methods come at a high computational cost as the size of the team grows and are unsuitable for online applications (see Fig. \ref{fig:computation_time}). On the other hand, inference on a trained neural network is inexpensive. Thus, as a solution to this scaling problem, we propose a supervised learning approach wherein a CNN learns to position mobile communication nodes according to solutions obtained from an optimization scheme similar to \cite{kim2005maximizing}.
\subsection{Communication Model}
\label{sec:org92bc834}
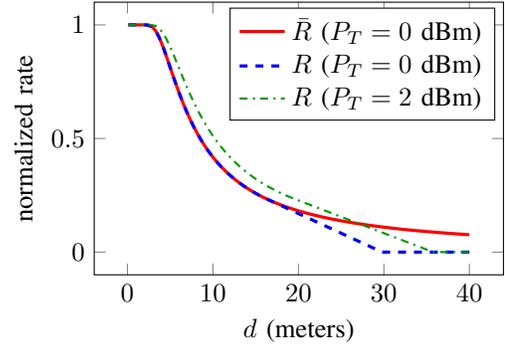
\begin{figure}
  \centering
  \begin{tikzpicture}
    \begin{axis}[
      height = 0.20\textwidth,
      width = 0.30\textwidth,
      scale only axis,
      xlabel = {$d$ (meters)},
      ylabel = {normalized rate},
    ]
    \addplot [color=red,very thick,solid] table [x=x,y=cm0,col sep=comma,] {figures/channel_model.csv};
    \addplot [color=blue,very thick,dashed] table [x=x,y=cm1,col sep=comma,] {figures/channel_model.csv};
    \addplot [color=green!60!black,thick,dashdotted] table [x=x,y=cm2,col sep=comma,] {figures/channel_model.csv};
    \legend{$\bar{R}$ ($P_T=0$ dBm), $R$ ($P_T=0$ dBm), $R$ ($P_T=2$ dBm)}
    \end{axis}
  \end{tikzpicture}
  \caption{\label{fig:channel_model}The non-vanishing wireless channel model in Eq. \eqref{eq:path_loss} and the piecewise smooth model of Eq. \eqref{eq:channel_model} with $n = 2.52$, $P_{N_0} = -70$ dBm, and $K=5.01\times 10^{-6}$ (-53 dB).}
\end{figure}

In order to reason about communication in a group of robots we must be able to predict the ability of pairs of agents to exchange information. In this work we use a function of the distance between two agents borrowed from probabilistic channel approaches that balances accuracy with model complexity by capturing the dominant fading characteristics of wireless channels \cite{fink2011communication}. Concretely, agents \(i\) and \(j\) located at \(x_i, x_j \in \mathbb{R}^2\), respectively, can communicate according to the following wireless channel normalized rate function:
\begin{align}
\bar{R}(d) = \text{erf} \left( \sqrt{\frac{P_R}{P_{N_0}}d^{-n}} \right)
\label{eq:path_loss}
\end{align}
where \(P_R\) is the received signal power computed from:
\begin{align}
P_R = P_TKd^{-n}.
\label{eq:rss}
\end{align}
In the above equations, \(P_T\) is the transmit power (mW), \(d=||x_i-x_j||\), \(K\) is a constant specific to the communication hardware used, \(P_{N_0}\) is the ambient noise at the receiver (mW), and \(n\) is the signal decay exponent \cite{fink2011communication}. Note that while Eqs. \eqref{eq:path_loss}, \eqref{eq:rss} assume absolute power (mW), it is often convenient to refer to various quantities in relative units of decibel-milliwatts (dBm). It is also useful to express Eq. \eqref{eq:path_loss} in terms of the robot positions as \(\bar{R}(x_i,x_j) = \bar{R}(||x_i-x_j||)\). Note that Eq. \eqref{eq:path_loss} approaches but does not reach zero for increasing \(d\), implying that two agents can communicate at arbitrarily large distances, albeit at very low rates (see Fig. \ref{fig:channel_model}). This behavior does not align with reality as a wireless channel will certainly cease to function at large distances as the signal becomes indistinguishable from noise. Thus, we wrap Eq. \eqref{eq:path_loss} in a piecewise smooth function that eventually drives the channel rate to zero:
\begin{align}
  R(d) =
    \begin{cases}
      \bar{R}(d) & d \leq d_t\\
      \left. \frac{\partial\bar{R}}{\partial d} \right\rvert_{d_t}(d-d_t) + R(d_t) & d_t < d \leq d_c\\
      0 & d_c < d
    \end{cases}
    \label{eq:channel_model}
\end{align}
where the transition distance \(d_t\) is found from a chosen, fixed cutoff rate and \(d_c\) is the cutoff distance at the zero rate crossing of the linear function. A plot of Eq. \eqref{eq:channel_model} can be seen in Fig. \ref{fig:channel_model}. In this work we use a default set of parameters for Eq. \eqref{eq:channel_model} (see Fig. \ref{fig:channel_model}) with a fixed cutoff rate but allow the transmit power to be varied from its default value of \(0\) dBm when necessary, which effectively extends the max range of the channel, \(d_c\).

\subsection{Dataset Generation}
\label{sec:org0c6abe7}
Our CNN learns to position communication agents from an expert. In this case, the expert is an optimization approach to maximizing the connectivity of a communication graph based on \cite{kim2005maximizing}:
\begin{subequations}
  \begin{align}
    \underset{\boldsymbol{x}_C^{k+1}}{\text{maximize}} \quad& \gamma\nonumber\\
    \text{subject to} \quad& P^TLP \succeq I\gamma\label{eq:lambda2_constraint}\\
    & L = \text{diag}(A\boldsymbol{1}) - A\nonumber\\
    & [A]_{ij} = R(x_i^k, x_j^k)+ \nabla_{x_i^{k+1}}R^T(x_i^{k+1} - x_i^k)\nonumber\\
    & \hspace{5em} + \nabla_{x_j^{k+1}}R^T(x_j^{k+1} - x_j^k)\label{eq:linear_channel_model}\\
    & |\boldsymbol{x}_C^{k+1} - \boldsymbol{x}_C^k|_1 \leq \Delta\label{eq:proximity_constraint}
  \end{align}
  \label{eq:connectivity_sdp}
\end{subequations}  
where \(L\) and \(A\) are the state dependent graph Laplacian and adjacency matrix, respectively, and \(P\) is an orthogonal basis spanning \(\boldsymbol{1}^\perp\). The tunable parameter \(\Delta\) along with the constraint in Eq. \eqref{eq:proximity_constraint} ensures that the optimization variables \(\boldsymbol{x}_C^{k+1}\) remain in a region around the current configuration \(\boldsymbol{x}_C^k\) where the linearized channel model in Eq. \eqref{eq:linear_channel_model} is valid. Eq. \eqref{eq:lambda2_constraint} forces the algebraic connectivity to be greater than or equal to \(\gamma\). Note that the optimization variables are the communication agent positions \(\boldsymbol{x}_C\) as they are the only agents that can be controlled. Problem \eqref{eq:connectivity_sdp} adjusts \(\boldsymbol{x}_C\) in order to maximize the algebraic connectivity of the team, \(\gamma\), and can be solved in an iterative fashion using an available SDP solver, converging to a local optimum given an initial feasible configuration. For more detail see \cite{kim2005maximizing}.

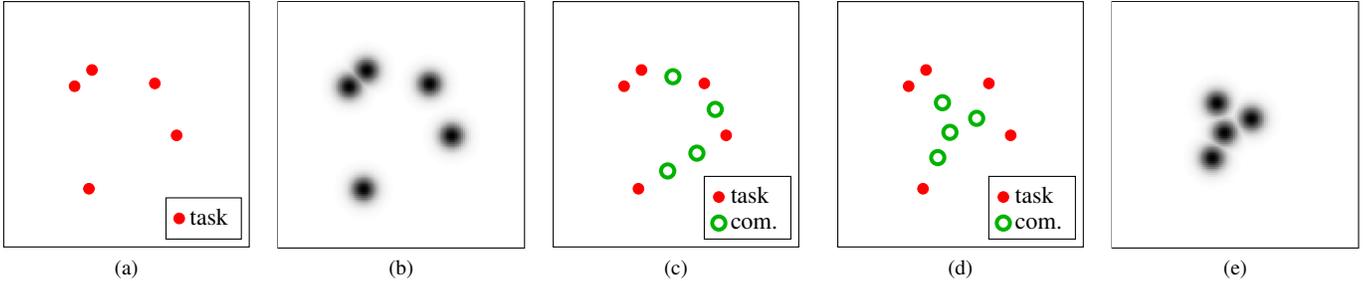
\begin{figure*}
  \centering
  \subfloat[][]{
    \begin{tikzpicture}
      \begin{axis}[legend style={cells={anchor=west}, font=\small}, legend pos=south east, height=0.18\textwidth, width=0.18\textwidth, ticks=none, scale only axis, xmin=-80, xmax=80, ymin=-80, ymax=80,]
      \addplot [red,only marks] coordinates {(-22.47852457, 35.40199216) (-33.81412171, 24.80760193) (-24.45805657,-42.03612643) ( 18.47205358, 26.64574924) ( 32.67891161, -7.23171668)};
      \legend{task}
      \end{axis}
    \end{tikzpicture}
    \label{fig:dataset_task_config}
  }
  \hfill
  \subfloat[][]{\frame{\includegraphics[width=0.18\textwidth]{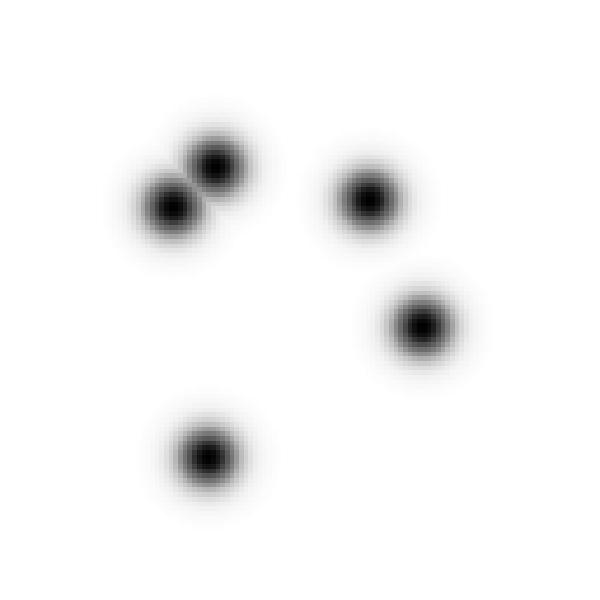}}\label{fig:dataset_task_image}}
  \hfill
  \subfloat[][]{
    \begin{tikzpicture}
      \begin{axis}[legend style={cells={anchor=west}, font=\small}, legend pos=south east, height=0.18\textwidth, width=0.18\textwidth, ticks=none, scale only axis, xmin=-80, xmax=80, ymin=-80, ymax=80,]
      \addplot [red,only marks] coordinates {(-22.47852457, 35.40199216) (-33.81412171, 24.80760193) (-24.45805657,-42.03612643) ( 18.47205358, 26.64574924) ( 32.67891161, -7.23171668)};
      \addplot [mygreen,only marks,mark=o,mark size=2.5,line width=1.5] coordinates {( 13.63325555,-18.8331866 ) ( -5.41240051,-30.43465652) ( -2.0032355 , 31.0238707 ) ( 25.5754826 ,  9.70701628)};
      \legend{task,\raisebox{-1.5px}{com.}}
      \end{axis}
    \end{tikzpicture}
    \label{fig:dataset_mst_config}
  }
  \hfill
  \subfloat[][]{
    \begin{tikzpicture}
      \begin{axis}[legend style={cells={anchor=west}, font=\small}, legend pos=south east, height=0.18\textwidth, width=0.18\textwidth, ticks=none, scale only axis, xmin=-80, xmax=80, ymin=-80, ymax=80,]
      \addplot [red,only marks] coordinates {(-22.47852457, 35.40199216) (-33.81412171, 24.80760193) (-24.45805657,-42.03612643) ( 18.47205358, 26.64574924) ( 32.67891161, -7.23171668)};
      \addplot [mygreen,only marks,mark=o,mark size=2.5,line width=1.5] coordinates {( -6.91169329, -5.26172446) (-14.8381976 ,-21.7396768 ) (-11.77748397, 14.07564251) ( 10.52228174,  3.86677216)};
      \legend{task,\raisebox{-1.5px}{com.}}
      \end{axis}
    \end{tikzpicture}
    \label{fig:dataset_final_config}
  }
  \hfill
  \subfloat[][]{\frame{\includegraphics[width=0.18\textwidth]{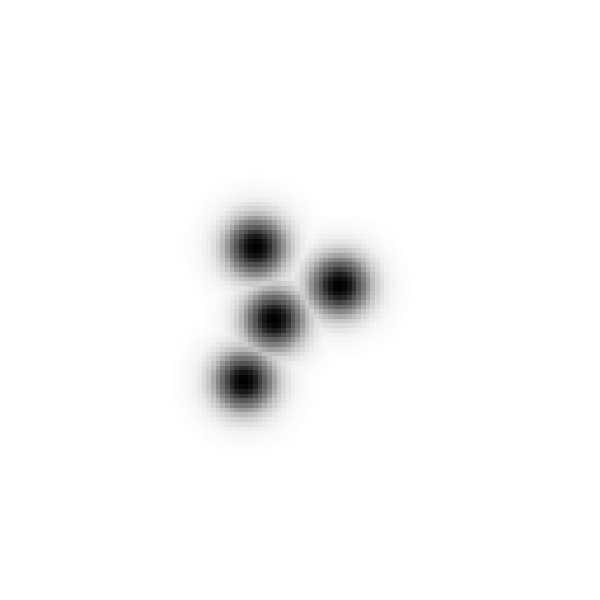}}\label{fig:dataset_comm_image}}
  \caption{\label{fig:dataset_image}(a) A randomly generated task team configuration; (b) The corresponding task team image; (c) The augmented MST team configuration derived from (a); (d) The locally optimal team configuration found by solving \eqref{eq:connectivity_sdp}; (e) The network team image corresponding to (d). Each image has been cropped from 256x256 to 128x128 pixels which represents a square area with a side length of 160 meters.}
\end{figure*}

We use the optimization expert in Eq. \eqref{eq:connectivity_sdp} to generate training samples consisting of pairs of images that capture the given task team and target communication team configurations. In order to ensure that spatial relationships between agents are consistent throughout the dataset, the metric distance each image pixel represents is fixed at 1.25 meters per pixel. Task team configurations \(\boldsymbol{x}_T\) are sampled from a uniform distribution (Fig. \ref{fig:dataset_task_config}) and marked in the image using a Gaussian kernel to avoid issues with sparsity (Fig. \ref{fig:dataset_task_image}). Next, a feasible initial network team configuration must be found to seed the optimization. To do this, a minimum spanning tree (MST) of the task team configuration is computed and communication agents are used to break up graph edges longer than \(d_c\) (Fig. \ref{fig:dataset_mst_config}). The corresponding communication graph can be readily found from the augmented MST by applying Eq. \eqref{eq:channel_model} to each pair of agents and keeping edges where the communication rate is greater than zero. The resulting graph is guaranteed to be connected since no edge in the augmented MST is greater than \(d_c\).

With this feasible initial solution, Problem \eqref{eq:connectivity_sdp} can be iteratively solved to adjust the network node locations into a locally optimal configuration (Fig. \ref{fig:dataset_final_config}). The resulting communication team configuration is marked in a separate image using the same Gaussian kernel (Fig. \ref{fig:dataset_comm_image}). Figs. \ref{fig:dataset_task_image} and \ref{fig:dataset_comm_image} constitute the input and target output images of the CNN, respectively. This entire data generation sequence is illustrated for one sample in Fig. \ref{fig:dataset_image}.
The richness of the training dataset is critical to the learned model's performance. In problems where training samples are scarce it is common to perform extensive data augmentation to increase the size of the dataset and ensure the samples capture sufficient angular diversity to account for the rotational sensitivity of CNNs. In our case, training samples are not only abundant but also randomly generated in a way that the dataset naturally encodes sufficient angular diversity for the CNN to learn the rotational symmetries of our problem. In light of this, we do not perform any explicit data augmentation.
\subsection{Learning Architecture}
\label{sec:orga21e708}
When it comes to learning connectivity maximizing configurations, choosing an appropriate model architecture is of paramount importance. Our choice of a CNN architecture might seem surprising. Why not use a learning model that accepts the position of task agents and produces communication agent positions? In fact, a CNN with an image as an input is a more appropriate representation because it leverages the symmetries of the connectivity problem. We can think of the communication team as filling in the gaps between the members of the task team, a goal that depends on the relative positions of agents and is invariant to their absolute positions. CNNs possess this very property making them a suitable choice for this problem. Furthermore, images are a natural way to represent spatial information \emph{without incurring penalties with scaling up to large numbers of agents}. Provided the image is sized to cover an adequate metric area, configurations with many agents can be readily represented alongside those with few. Once in image form, all inputs to the CNN are processed equally meaning there is no performance difference between a team consisting of 4 agents or 20. This is not the case with the optimization in Eq. \ref{eq:connectivity_sdp}, which becomes prohibitively slow for large teams (see Fig. \ref{fig:computation_time}).

Considering the input and output of our network is an image, we employ an autoencoder (AE) like architecture comprised entirely of convolutional layers which we refer to as ConvAE. While typical AEs use fully connected layers or a probability distribution at the information bottleneck between the encoder and decoder, we found that using convolutional layers instead resulted in better generalization performance, especially as the size of the task team increased. As a side effect, ConvAE can operate on arbitrarily sized input images provided they pass cleanly through the network (for our model compatible image resolutions are given by \((N+4)\cdot 2^6\) for \(N\in\mathbb{Z}\geq 0\)). In other words, we never run out of image space to represent teams with many agents spread over large areas. Interestingly, for the smallest image resolution that the CNN can process, the convolutions at the bottleneck are effectively fully connected neurons. The encoder transforms the 256x256 input image into a volume with dimension 4x4xF, where F is the number of filters. Immediately after, the first convolutional transpose layer of the decoder takes this 4x4xF volume and processes it with a 4x4 kernel. Notice how the entire transformed input volume fits cleanly into the 4x4 convolutional transpose filter so that every extracted feature factors into each filter’s output, exactly like a fully connected layer. The utility of autoencoders lies in the information bottleneck, which forces the network to learn a lower dimensional representation of the input data in order to faithfully replicate the output. In our case, the information bottleneck at the deepest convolutional layer causes the network to glean salient geometric features from the input configuration image that inform the placement of network agents in the output image. The details of our architecture are shown in Table \ref{tab:cnn_architecture}.

\begin{table}[htbp]
\caption{\label{tab:cnn_architecture}CNN architecture: network proceeds sequentially from top (input) to bottom (output). Each Conv2D / ConvTranspose2D layer is padded, uses a stride of 2, and contains 128 input and output channels.}
\centering
\small
\begin{tabular}{llrrl}
model & layer type & count & kernel & activation\\
\hline
encoder & Conv2D & 2 & 8x8 & LeakyReLU\\
encoder & Conv2D & 4 & 4x4 & LeakyReLU\\
decoder & ConvTranspose2D & 4 & 4x4 & ReLU\\
decoder & ConvTranspose2D & 2 & 8x8 & ReLU\\
\end{tabular}
\end{table}

\section{Results}
\label{sec:org85ae6b9}
\begin{figure}
  \centering
  \subfloat[]{\frame{\includegraphics[width=0.155\textwidth]{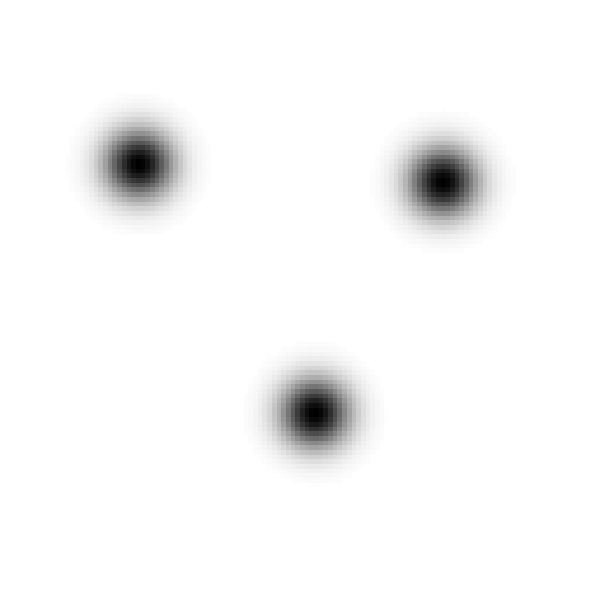}}\label{fig:cnn_test_example_input}}
  \hfill
  \subfloat[]{\frame{\includegraphics[width=0.155\textwidth]{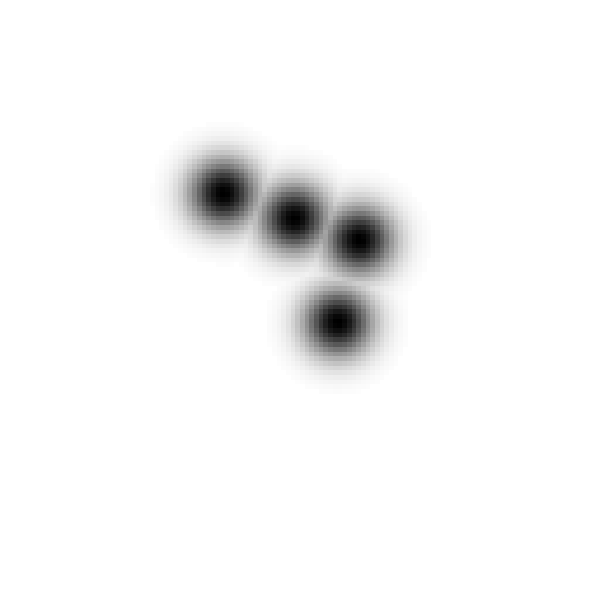}}\label{fig:cnn_test_example_output}}
  \hfill
  \subfloat[]{\frame{\includegraphics[width=0.155\textwidth]{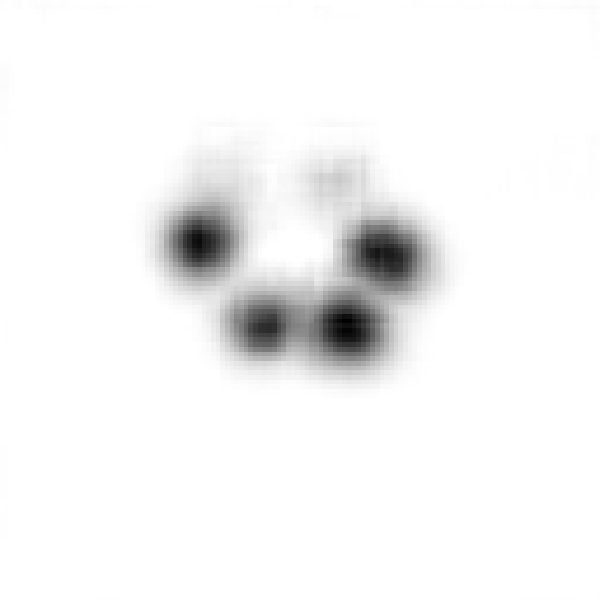}}\label{fig:cnn_test_example_model}}
  \caption{\label{fig:cnn_test_example}results for one test sample showing (a) the input image to the CNN; (b) the expected output; and (c) the CNN output. Each image has been cropped from 256x256 to 128x128 for clarity.}
\end{figure}

Our ConvAE model was trained for 14 epochs on a dataset comprised of 595k images generated using the process described in Section \ref{sec:org0c6abe7}, with task teams comprised of 2-6 agents. Adam was used as the optimizer with a learning rate of \(10^{-4}\), a batch size of 4, and mean-squared-error as the loss function \footnote{code and multimedia available at: \url{www.danmox.com/projects/convae.html}}. The output of the CNN for one test sample can be seen in Fig. \ref{fig:cnn_test_example}. It is immediately apparent that the CNN output in Fig. \ref{fig:cnn_test_example_model} differs from that of the optimization in Fig. \ref{fig:cnn_test_example_output}; what may not be so obvious is that the CNN yields a valid configuration that connects the task agents shown in Fig. \ref{fig:cnn_test_example_input}. This outcome is not unexpected. The optimization produces \emph{locally optimal} solutions and for a given task team configuration there may be many distinct local optimums. Clearly the performance of the CNN must be judged by its ability to produce connected configurations and not the relative closeness of the output image to the reference image in a mean-squared-error sense.

In order to compute algebraic connectivity we must be able to extract relay node positions in \(\mathbb{R}^2\) from the output images of the CNN. While the peaks in Fig. \ref{fig:cnn_test_example_model} are distinct and easy to pick out, the CNN images are not always so clear (see Fig. \ref{fig:circle_test_c}). To overcome this issue we take a coverage perspective, interpreting the output image as a distribution the communication agents must assume. First, we determine the number of agents to deploy by utilizing an adaptive thresholding scheme to pick out peaks in the intensity image. Then, we employ Lloyd's algorithm for the specified number of agents to find a configuration that achieves locally optimal coverage of the intensity distribution \cite{cortes2004coverage}. The CNN may output redundant agents, especially when the task team configuration is symmetric (see Figs. \ref{fig:circle_test_d}, \ref{fig:circle_test_f}). We prune these extra agents by extracting a minimal connected sub-graph. Finally, the algebraic connectivity can readily be found using the known positions of the task agents used to generate the input image.

Algebraic connectivity alone does not tell the whole story. Since our channel model in Eq. \eqref{eq:channel_model} is truncated, there is a hard cutoff at \(d_c\) beyond which it is assumed no communication is possible. Any configuration relying on an edge greater than \(d_c\) has an algebraic connectivity of zero (i.e. is disconnected). However, for performance evaluation we want to know how close a configuration is to being connected (i.e. distinguish between one configuration relying on an edge one centimeter beyond \(d_c\) and another tied together with an edge many meters greater than \(d_c\)). Thus, as a more informative criterion, we consider the transmit power \(P_T\) required to achieve a connected configuration. For the optimization this value is guaranteed to be \(0\) dBm. For the CNN we extract the configuration as described above and check connectivity with \(0\) dBm, increasing it if necessary until connectivity is established. In this way, the performance of our system is quantified by the amount of power required to connect the network as compared to the optimization.

The following Sections \ref{sec:orgacdd2f9} and \ref{sec:org1a537c4} provide qualitative results of our system on canonical line and ring topologies. Afterwards, we present ConvAE's performance over an extensive test dataset in Section \ref{sec:org403a77f}, show its ability to generalize to larger teams not seen during training in Section \ref{sec:orgf06b5e4}, detail how our approach scales in Section \ref{sec:org9641884}, and finally demonstrate how our system can be deployed in real, dynamic teams of robots in Section \ref{sec:orge379529}.

\definecolor{mygreen}{rgb}{0,0.7,0}
\definecolor{myyellow}{rgb}{0.7,0.7,0}
\definecolor{myblue}{rgb}{0.0,0.3,1.0}
\newcommand{\figcaption}{Dimensions are in meters. Relay agents are shown as x's for the CNN and o's for the optimization (refered to as opt.), with the number deployed by each method shown in the legend, and the background image combines the input and output of the CNN. The title format is: method (algebraic connectivity, transmit power in dBm)}
\newcommand{\linecircleimgwidth}{0.19}

\subsection{Line Test}
\label{sec:orgacdd2f9}
\begin{figure}
  \centering
  \subfloat[][]{
    \begin{tikzpicture}
      \begin{axis}[title={opt. (0.11, 0) | CNN (0.11, 0)}, title style={at={(0.50,0.8)}, font=\scriptsize}, legend style={cells={anchor=west}, font=\scriptsize}, legend columns=4, legend pos=south west, height=\linecircleimgwidth\textwidth, width=\linecircleimgwidth\textwidth, ticks=none, scale only axis, xmin=-80, xmax=80, ymin=-80, ymax=80,enlargelimits=false,axis on top,]
      \addplot graphics [xmin=-80,xmax=80,ymin=-80,ymax=80] {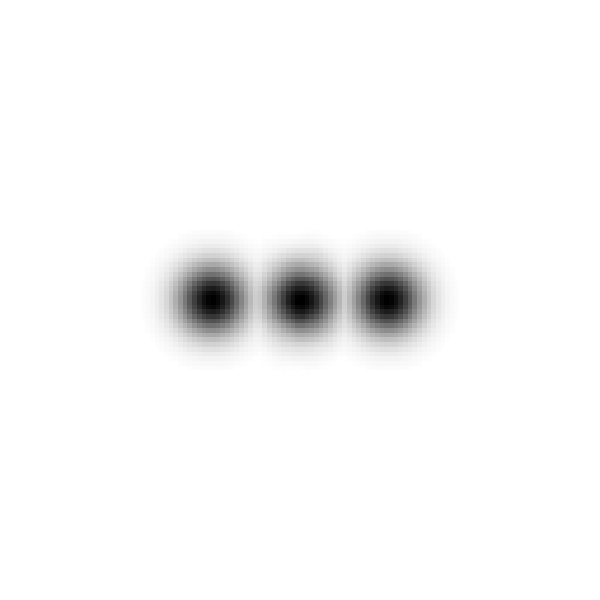};
      \addplot [red,only marks,mark size=1.7] coordinates {(-23.39, 0.0) (23.39, 0.0)};
      \addplot [mygreen,only marks,mark=o,mark size=2.5,line width=1.5] coordinates {(0.0, 0.0)};
      \addplot [myblue,only marks,mark=x,mark size=3.5,line width=1.5] coordinates {(0.55, -0.16)};
      \legend{,task,opt(1),CNN(1)}
      \end{axis}
    \end{tikzpicture}
    \label{fig:line_test_a}
  }
  \hfil
  \subfloat[][]{
    \begin{tikzpicture}
      \begin{axis}[title={opt. (0.067, 0) | CNN (0.067, 0)}, title style={at={(0.50,0.8)}, font=\scriptsize}, legend style={cells={anchor=west}, font=\scriptsize}, legend columns=4, legend pos=south west, height=\linecircleimgwidth\textwidth, width=\linecircleimgwidth\textwidth, ticks=none, scale only axis, xmin=-80, xmax=80, ymin=-80, ymax=80,enlargelimits=false,axis on top,]
      \addplot graphics [xmin=-80,xmax=80,ymin=-80,ymax=80] {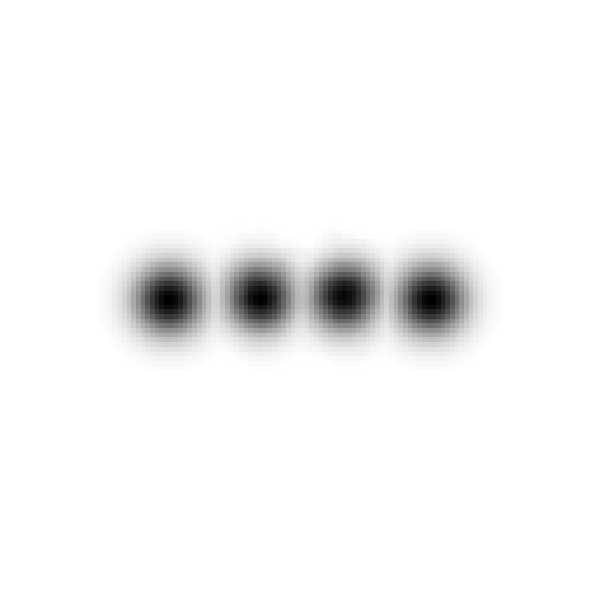};
      \addplot [red,only marks,mark size=1.7] coordinates {(-34.91, 0.0) (34.91, 0.0)};
      \addplot [mygreen,only marks,mark=o,mark size=2.5,line width=1.5] coordinates {(-10.98, 0.0) (10.98, 0.0)};
      \addplot [myblue,only marks,mark=x,mark size=3.5,line width=1.5] coordinates {(-10.59, 0.6286) (11.776, 0.8856)};
      \legend{,task,opt(2),CNN(2)}
      \end{axis}
    \end{tikzpicture}
    \label{fig:line_test_b}
  }\\
  \subfloat[][]{
    \begin{tikzpicture}
      \begin{axis}[title={opt. (0.053, 0) | CNN (0.052, 0)}, title style={at={(0.50,0.8)}, font=\scriptsize}, legend style={cells={anchor=west}, font=\scriptsize}, legend columns=4, legend pos=south west, height=\linecircleimgwidth\textwidth, width=\linecircleimgwidth\textwidth, ticks=none, scale only axis, xmin=-80, xmax=80, ymin=-80, ymax=80,enlargelimits=false,axis on top,]
      \addplot graphics [xmin=-80,xmax=80,ymin=-80,ymax=80] {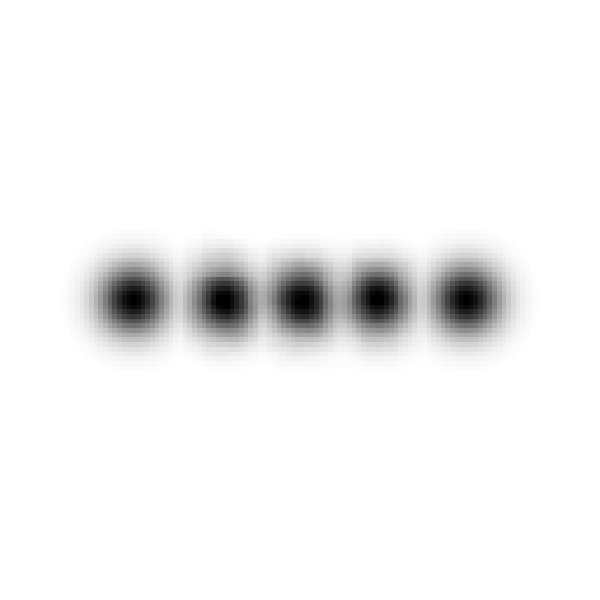};
      \addplot [red,only marks,mark size=1.7] coordinates {(-44.41, 0.00) (44.41, 0.00)};
      \addplot [mygreen,only marks,mark=o,mark size=2.5,line width=1.5] coordinates {(-20.65, 0.00) (0.04, 0.00) (20.73, 0.00)};
      \addplot [myblue,only marks,mark=x,mark size=3.5,line width=1.5] coordinates {(-20.29, -0.45) (0.54, -0.34) (20.67, 0.21)};
      \legend{,task,opt(3),CNN(3)}
      \end{axis}
    \end{tikzpicture}
    \label{fig:line_test_c}
  }
  \hfil
  \subfloat[][]{
    \begin{tikzpicture}
      \begin{axis}[title={opt. (0.029, 0) | CNN (0.028, 0)}, title style={at={(0.50,0.8)}, font=\scriptsize}, legend style={cells={anchor=west}, font=\scriptsize}, legend columns=4, legend pos=south west, height=\linecircleimgwidth\textwidth, width=\linecircleimgwidth\textwidth, ticks=none, scale only axis, xmin=-80, xmax=80, ymin=-80, ymax=80,enlargelimits=false,axis on top,]
      \addplot graphics [xmin=-80,xmax=80,ymin=-80,ymax=80] {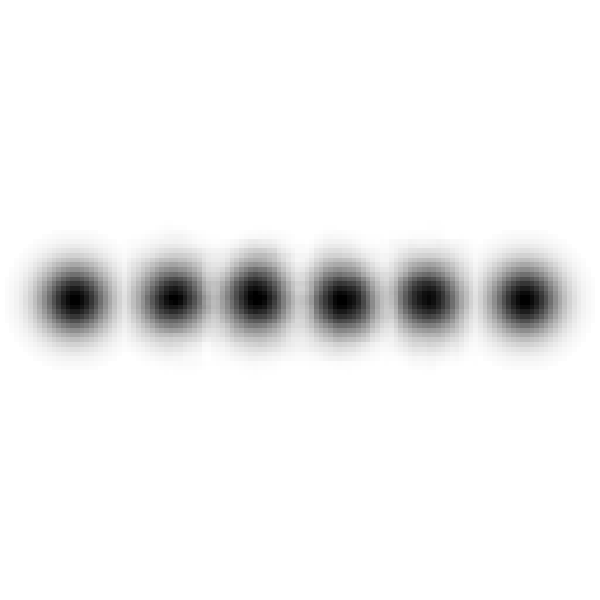};
      \addplot [red,only marks,mark size=1.7] coordinates {(-60.00, 0.00) (60.00, 0.00)};
      \addplot [mygreen,only marks,mark=o,mark size=2.5,line width=1.5] coordinates {(-34.36, 0.00) (-11.19, 0.00) (11.20, 0.00) (34.35, 0.00)};
      \addplot [myblue,only marks,mark=x,mark size=3.5,line width=1.5] coordinates {(-33.71, 0.51) (-11.70, 0.59) (11.61, -0.34) (34.17, 0.41)};
      \legend{,task,opt(3),CNN(3)}
      \end{axis}
    \end{tikzpicture}
    \label{fig:line_test_d}
  }
  \caption{\label{fig:line_test}Line scenario results progressing from (a) to (d). \figcaption. The images have been cropped from 256x256 to 128x128 for clarity, covering a 160x160m area.}
\end{figure}

Perhaps the simplest test of the CNN's ability to produce connected configurations is a line test. Two agents starting relatively close together progressively move away from one another, necessitating the formation of a chain of communication relay nodes in between. Snapshots of the CNN and optimization results can seen in Fig. \ref{fig:line_test}.

Across the snapshots, the configurations and corresponding connectivity values produced by the CNN and the optimization are very similar. This is not surprising as the CNN was trained on 170k images of 2 agent task teams of varying density and orientation, providing ample opportunity for the model to learn line topologies. In all cases, the CNN was able to produce connected configurations with the same number of agents as the optimization without needing to vary transmit power from its default value of \(0\) dBm.
\subsection{Circle Test}
\label{sec:org1a537c4}
\begin{figure}
  \centering
  \subfloat[][]{
    \begin{tikzpicture}
      \begin{axis}[title={opt. (0.41, 0) | CNN (0.13, 0)}, title style={at={(0.50,0.8)}, font=\scriptsize}, legend style={cells={anchor=west}, font=\scriptsize}, legend columns=4, legend pos=south west, height=\linecircleimgwidth\textwidth, width=\linecircleimgwidth\textwidth, ticks=none, scale only axis, xmin=-80, xmax=80, ymin=-80, ymax=80,enlargelimits=false,axis on top,]
      \addplot graphics [xmin=-80,xmax=80,ymin=-80,ymax=80] {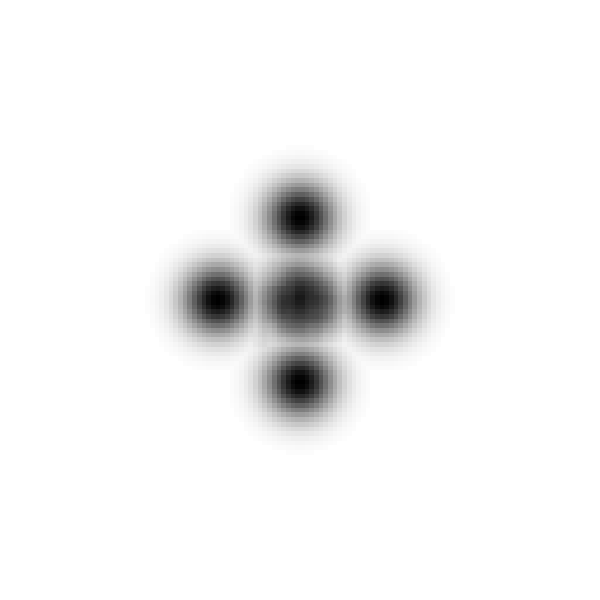};
      \addplot [red,only marks,mark size=1.7] coordinates {(21.92, 0.00) (0.00, 21.92) (-21.92, 0.00) (-0.00, -21.92)};
      \addplot [mygreen,only marks,mark=o,mark size=2.5,line width=1.5] coordinates {(0.02, 0.00) (0.00, 0.01) (-0.02, -0.02)};
      \addplot [myblue,only marks,mark=x,mark size=3.5,line width=1.5] coordinates {(0.16, -0.17)};
      \legend{,task,opt(3),CNN(1)}
      \end{axis}
    \end{tikzpicture}
    \label{fig:circle_test_a}
  }
  \hfil
  \subfloat[][]{
    \begin{tikzpicture}
      \begin{axis}[title={opt. (0.173, 0) | CNN (0.076, 0)}, title style={at={(0.50,0.8)}, font=\scriptsize}, legend style={cells={anchor=west}, font=\scriptsize}, legend columns=4, legend pos=south west, height=\linecircleimgwidth\textwidth, width=\linecircleimgwidth\textwidth, ticks=none, scale only axis, xmin=-80, xmax=80, ymin=-80, ymax=80,enlargelimits=false,axis on top,]
      \addplot graphics [xmin=-80,xmax=80,ymin=-80,ymax=80] {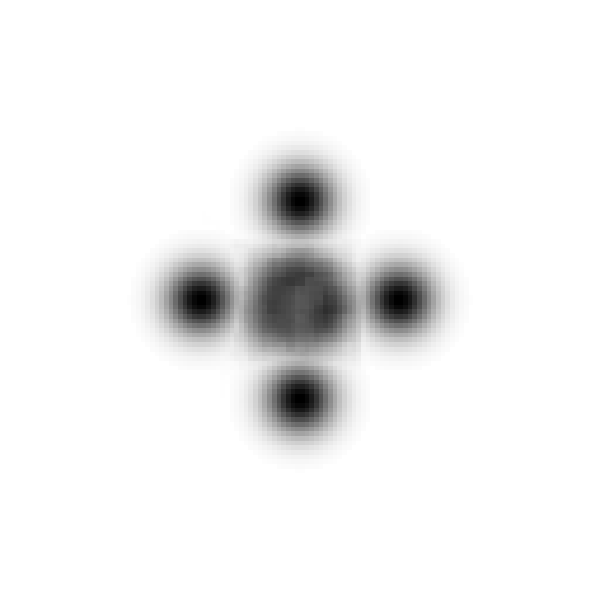};
      \addplot [red,only marks,mark size=1.7] coordinates {(26.44, 0.00) (0.00, 26.44) (-26.44, 0.00) (-0.00, -26.44)};
      \addplot [mygreen,only marks,mark=o,mark size=2.5,line width=1.5] coordinates {(5.71, 5.71) (1.30, -7.20) (-7.21, 1.28)};
      \addplot [myblue,only marks,mark=x,mark size=3.5,line width=1.5] coordinates {(-5.54, 2.88) (6.66, -3.22)};
      \legend{,task,opt(3),CNN(2)}
      \end{axis}
    \end{tikzpicture}
    \label{fig:circle_test_b}
  }\\
  \subfloat[][]{
    \begin{tikzpicture}
      \begin{axis}[title={opt. (0.026, 0) | CNN (0.085, 0)}, title style={at={(0.50,0.8)}, font=\scriptsize}, legend style={cells={anchor=west}, font=\scriptsize}, legend columns=4, legend pos=south west, height=\linecircleimgwidth\textwidth, width=\linecircleimgwidth\textwidth, ticks=none, scale only axis, xmin=-80, xmax=80, ymin=-80, ymax=80,enlargelimits=false,axis on top,]
      \addplot graphics [xmin=-80,xmax=80,ymin=-80,ymax=80] {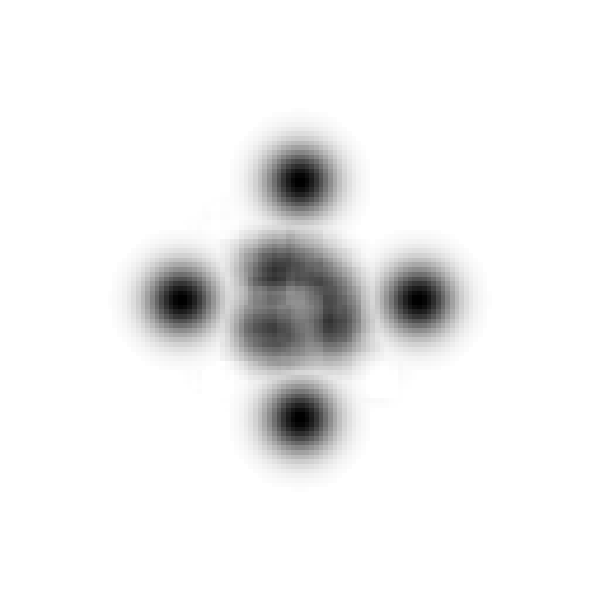};
      \addplot [red,only marks,mark size=1.7] coordinates {(31.60, 0.00) (0.00, 31.60) (-31.60, 0.00) (-0.00, -31.60)};
      \addplot [mygreen,only marks,mark=o,mark size=2.5,line width=1.5] coordinates {(14.41, 17.17) (-15.76, 15.84) (-17.18, -14.42)};
      \addplot [myblue,only marks,mark=x,mark size=3.5,line width=1.5] coordinates {(-7.63, -8.60) (-4.36, 7.76) (8.57, -4.31)};
      \legend{,task,opt(3),CNN(3)}
      \end{axis}
    \end{tikzpicture}
    \label{fig:circle_test_c}
  }
  \hfil
  \subfloat[][]{
    \begin{tikzpicture}
      \begin{axis}[title={opt. (0.017, 0) | CNN (0.016, 0)}, title style={at={(0.50,0.8)}, font=\scriptsize}, legend style={cells={anchor=west}, font=\scriptsize}, legend columns=4, legend pos=south west, height=\linecircleimgwidth\textwidth, width=\linecircleimgwidth\textwidth, ticks=none, scale only axis, xmin=-80, xmax=80, ymin=-80, ymax=80,enlargelimits=false,axis on top,]
      \addplot graphics [xmin=-80,xmax=80,ymin=-80,ymax=80] {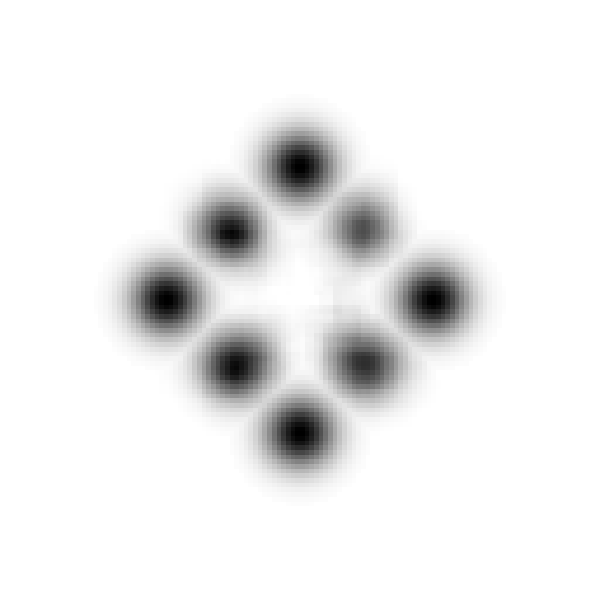};
      \addplot [red,only marks,mark size=1.7] coordinates {(35.47, 0.00) (0.00, 35.47) (-35.47, 0.00) (-0.00, -35.47)};
      \addplot [mygreen,only marks,mark=o,mark size=2.5,line width=1.5] coordinates {(17.73, 17.75) (18.65, -16.82) (-16.83, 18.64)};
      \addplot [myblue,only marks,mark=x,mark size=3.5,line width=1.5] coordinates {(-18.39, 18.30) (-16.88, -17.65) (16.72, 19.14)};
      \legend{,task,opt(3),CNN(3)}
      \end{axis}
    \end{tikzpicture}
    \label{fig:circle_test_d}
  }\\
  \subfloat[][]{
    \begin{tikzpicture}
      \begin{axis}[title={opt. (0.073, 0) | CNN (0.051, 0)}, title style={at={(0.50,0.8)}, font=\scriptsize}, legend style={cells={anchor=west}, font=\scriptsize}, legend columns=4, legend pos=south west, height=\linecircleimgwidth\textwidth, width=\linecircleimgwidth\textwidth, ticks=none, scale only axis, xmin=-80, xmax=80, ymin=-80, ymax=80,enlargelimits=false,axis on top,]
      \addplot graphics [xmin=-80,xmax=80,ymin=-80,ymax=80] {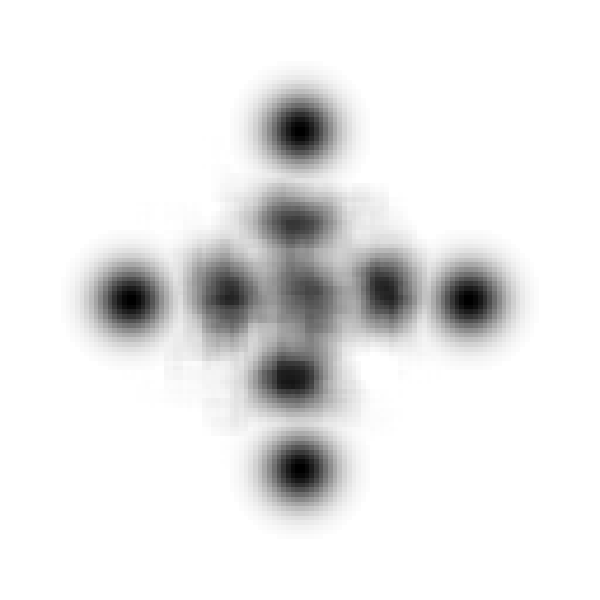};
      \addplot [red,only marks,mark size=1.7] coordinates {(45.16, 0.00) (0.00, 45.16) (-45.16, 0.00) (-0.00, -45.16)};
      \addplot [mygreen,only marks,mark=o,mark size=2.5,line width=1.5] coordinates {(22.74, -0.16) (0.76, -0.07) (-1.78, 22.17) (-22.08, 1.83) (-0.05, -0.58) (0.17, -22.73)};
      \addplot [myblue,only marks,mark=x,mark size=3.5,line width=1.5] coordinates {(-19.15, 0.80) (-2.06, -20.48) (-1.57, 19.60) (1.74, 0.86) (22.87, 2.16)};
      \legend{,task,opt(6),CNN(5)}
      \end{axis}
    \end{tikzpicture}
    \label{fig:circle_test_e}
  }
  \hfil
  \subfloat[][]{
    \begin{tikzpicture}
      \begin{axis}[title={opt. (0.009, 0) | CNN (0.008, 0)}, title style={at={(0.50,0.8)}, font=\scriptsize}, legend style={cells={anchor=west}, font=\scriptsize}, legend pos=south west, height=\linecircleimgwidth\textwidth, width=\linecircleimgwidth\textwidth, ticks=none, scale only axis, xmin=-80, xmax=80, ymin=-80, ymax=80,enlargelimits=false,axis on top,]
      \addplot graphics [xmin=-80,xmax=80,ymin=-80,ymax=80] {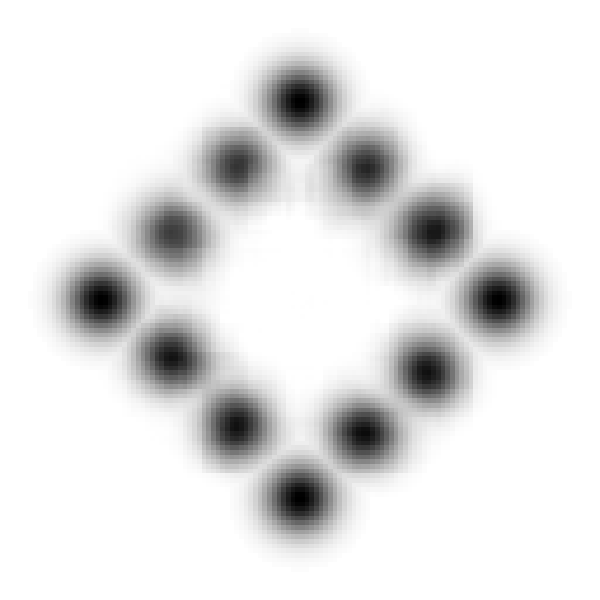};
      \addplot [red,only marks,mark size=1.7] coordinates {(52.90, 0.00) (0.00, 52.90) (-52.90, 0.00) (-0.00, -52.90)};
      \addplot [mygreen,only marks,mark=o,mark size=2.5,line width=1.5] coordinates {(35.19, 17.65) (17.68, 35.20) (36.73, -16.24) (19.05, -33.86) (-16.09, 36.74) (-33.79, 19.05)};
      \addplot [myblue,only marks,mark=x,mark size=3.5,line width=1.5] coordinates {(-33.82, 17.58) (-16.76, 35.82) (16.99, -35.71) (17.66, 35.15) (34.15, -19.22) (35.56, 18.37)};
      \legend{,,opt(6),CNN(6)}
      \end{axis}
    \end{tikzpicture}
    \label{fig:circle_test_f}
  }
  \caption{\label{fig:circle_test}Circle scenario results progressing from (a) to (f). \figcaption. The images have been cropped from 256x256 to 128x128 for clarity, covering a 160x160m area.}
\end{figure}

A harder test involves a group of task agents distributed on the perimeter of an expanding circle. The job of the optimization and CNN is to effectively deploy communication agents so that the task agents remain connected. Results for this circle test can be seen in Fig. \ref{fig:circle_test}.

Our CNN-based approach also performs well in the circle test when compared with the optimization. There are a few interesting results to note. First, because there is no explicit labeling in the images, the CNN only learns to paint one blob in cases where agent positions overlap. This can be seen in Fig. \ref{fig:circle_test_a}, \ref{fig:circle_test_e} where the optimization placed multiple overlapping agents while the CNN used fewer to connect the network.

Another significant takeaway from the circle test is that our CNN can outperform the optimization that it learned from, as seen in Fig. \ref{fig:circle_test_c}. Since algebraic connectivity decreases when links between agents are severed, the optimization will only ever maintain or increase the edges in the network and is thus confined to a local optimum that depends on the topology of the initial configuration. However, there are likely many different locally optimal configurations represented in the training data as randomly sampled task agent teams that are similar in shape may have different solutions from the optimization (see Figs. \ref{fig:circle_test_c} and \ref{fig:circle_test_d} where it yielded considerably different local optimums despite slight changes in the input configuration). Additionally, subsets of a given task agent configuration are likely to appear similar to other training samples. As a result, the CNN may output a configuration with a different topology than the optimization (see also Fig. \ref{fig:circle_test_c}).
\subsection{Dataset Statistics}
\label{sec:org403a77f}
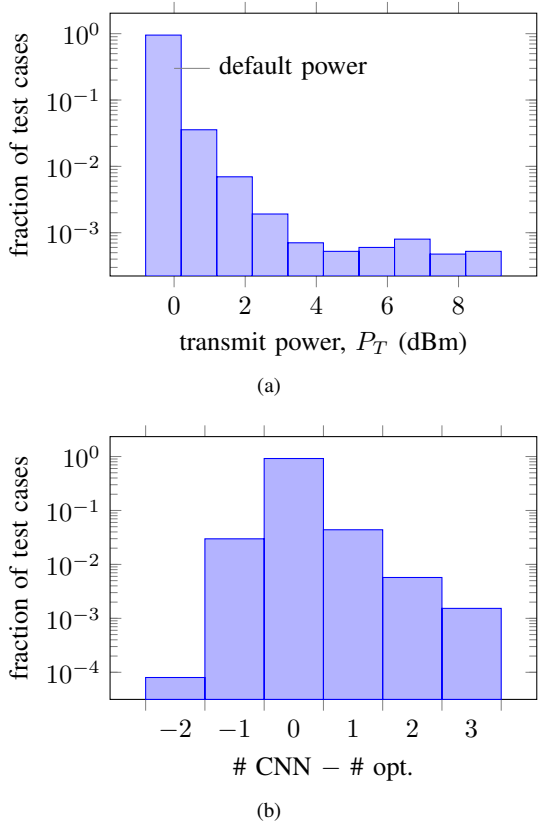
\begin{figure}
  \centering
  \subfloat[]{
    \begin{tikzpicture}
      \begin{semilogyaxis}[
        log origin y=infty,
        width = 0.40\textwidth,
        height = 0.28\textwidth,
        xlabel = {transmit power, $P_T$ (dBm)},
        xtick distance=2.0,
        xtickmax = 8,
        xtick align=outside,
        ylabel = {fraction of test cases},
        ]
        \addplot+ [ybar interval, mark=none, fill=blue!25, hist={data=x,data min=-0.8,data max=9.2,bins=10,density}] file {figures/transmit_power_2to6.csv};
        \node [coordinate,pin=right:{default power}] at (axis cs:0,0.3) {};
      \end{semilogyaxis}
    \end{tikzpicture}
    \label{fig:power_hist}
  }\\
  \subfloat[]{
    \begin{tikzpicture}
      \begin{semilogyaxis}[
        log origin y=infty,
        ybar interval,
        width = 0.40\textwidth,
        height = 0.28\textwidth,
        xlabel = {\# CNN $-$ \# opt.},
        ylabel = {fraction of test cases},
        ytick distance=10^1,
        grid=none,
        ]
        \addplot+ [hist={data=x,data min=-2,data max=4,bins=6,density}] file {figures/agent_diffs_2to6.csv};
      \end{semilogyaxis}
    \end{tikzpicture}
    \label{fig:agent_count_hist}
  }
  \caption{\label{fig:dataset_stats}Histograms of a) transmit power used by the CNN to achieve connectivity and b) the difference between the number of agents deployed by the CNN and optimization.}
\end{figure}

The line and circle tests provide a sense of how the CNN functions in two canonical scenarios. In this section we provide a more rigorous evaluation by computing the performance of ConvAE over a test dataset of 105k images generated in the same way as the training dataset with task teams of 2-6 agents. The results can be seen in Fig. \ref{fig:dataset_stats}.

As mention in the beginning of Section \ref{sec:org85ae6b9}, the CNN increases transmit power only when the configuration is not connected at the default value of \(0\) dBm. As can be seen in Fig. \ref{fig:power_hist}, the CNN produces connected configurations at \(0\) dBm (indicated by the leftmost bar) a vast majority of the time and in most other cases requires very little additional power. On average our CNN required \(0.05\) dBm of transmit power with a variance of \(+0.438\) dBm to produce connected configurations.

Fig. \ref{fig:agent_count_hist} shows a histogram comparing the number of agents deployed by the CNN and optimization. In cases where the CNN required more transmit power it was often due to deploying fewer agents. Thus to make this comparison fair, we restrict ourselves to test samples where the CNN used the default transmit power (100k out of the 105k cases) and show the number of agents deployed by the optimization subtracted from the number used by the CNN. Negative values indicate instances where the CNN used fewer agents to connect the network and positive values those where the CNN deployed more. In most cases the number of agents deployed by each method is within \(\pm1\) of each other showing that our approach achieves connectivity with roughly the same number of communication nodes as the optimization: \(0.031\pm 0.344\) more agents on average. In some cases, the CNN deploys more agents than required, typically for configurations near a topology change due to ambiguities in the training data. As a result, there is a slight asymmetry in the distribution in Fig. \ref{fig:agent_count_hist}; however, we note that the y-axis is a log scale and these instances represent a very small fraction of the 105k test cases.
\subsection{Generalization}
\label{sec:orgf06b5e4}
Our CNN-based approach performs well on test samples similar to ones seen during training. However, in practice it can be cumbersome to generate new data and retrain the network for every team size the CNN might encounter. Furthermore, even generating meaningful training data offline with the optimization becomes increasingly difficult as the number of agents increases. Given the symmetries of the problem, we expect our CNN to have learned something about connecting teams of 2-6 agents that would apply to larger teams.

\begin{figure}
  \centering
  \subfloat[]{
    \begin{tikzpicture}
      \tikzstyle{every node}=[font=\small]
      \begin{axis}[
        small,
        log origin y=infty,
        width = 0.25\textwidth,
        height = 0.22\textwidth,
        xlabel = {$P_T$ (dBm)},
        xtick distance=2.0,
        xtickmax = 8,
        xtick align=outside,
        ylabel = {fraction of test cases},
        ]
        \addplot+ [ybar interval, mark=none, fill=blue!25, hist={data=x,data min=-0.8,data max=9.2,bins=10,density}] file {figures/transmit_power_8and12.csv};
        \node [coordinate,pin={[align=left]0:{default\\power}}] at (axis cs:0,0.5) {};
      \end{axis}
    \end{tikzpicture}
    \label{fig:power_hist_812}
  }
  \subfloat[]{
    \begin{tikzpicture}
      \begin{axis}[
        small,
        log origin y=infty,
        ybar interval,
        width = 0.28\textwidth,
        height = 0.22\textwidth,
        xlabel = {\# CNN $-$ \# opt.},
        grid=none,
        ]
        \addplot+ [hist={data=x,data min=-2,data max=6,bins=8,density}] file {figures/agent_diffs_8and12.csv};
      \end{axis}
    \end{tikzpicture}
    \label{fig:agent_count_hist_812}
  }
  \caption{\label{fig:dataset_stats_812}Histograms of a) transmit power used by the CNN to achieve connectivity and b) the difference between the number of agents deployed by the CNN and optimization.}
\end{figure}
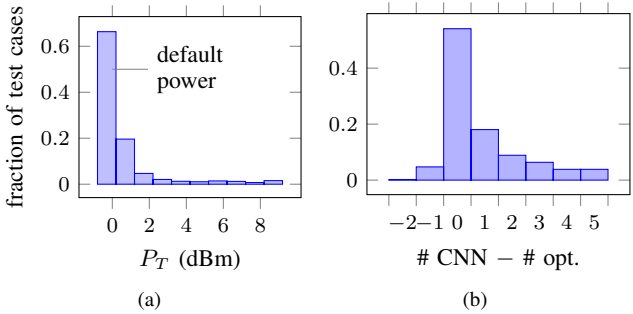

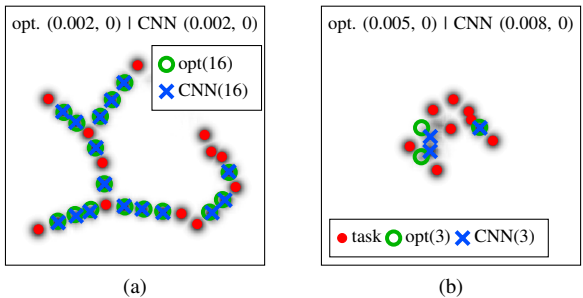
\begin{figure}
  \centering
  \subfloat[][]{
    \begin{tikzpicture}
      \begin{axis}[title={opt. (0.002, 0) | CNN (0.002, 0)}, title style={at={(0.50,0.8)}, font=\scriptsize}, legend style={at={(0.99,0.85)}, cells={anchor=west}, font=\scriptsize}, height=\linecircleimgwidth\textwidth, width=\linecircleimgwidth\textwidth, ticks=none, scale only axis, xmin=-160, xmax=160, ymin=-160, ymax=160,enlargelimits=false,axis on top,]
      \addplot graphics [xmin=-160,xmax=160,ymin=-160,ymax=160] {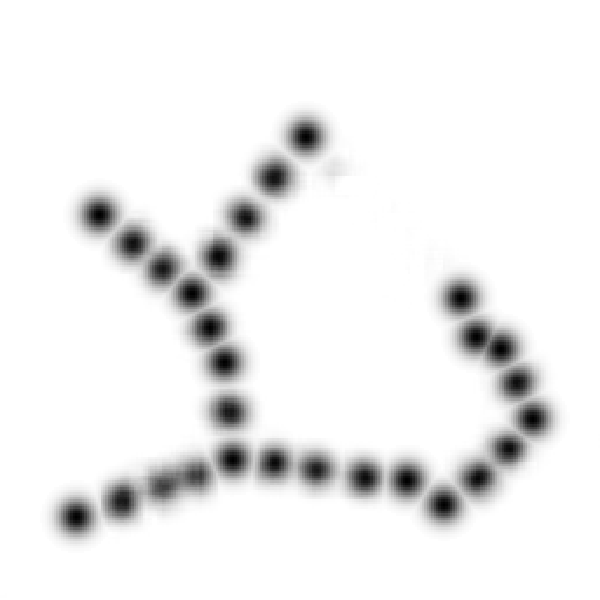};
      \addplot [red,only marks,mark size=1.7] coordinates {(3.22, 86.97) (76.75, -109.17) (107.15, -26.09) (57.25, -96.18) (-106.92, 45.39) (124.13, -63.43) (94.14, -19.73) (85.91, 0.95) (-35.84, -85.21) (-57.52, 3.52) (-119.20, -115.74) (-40.19, -33.37)};
      \addplot [mygreen,only marks,mark=o,mark size=2.5,line width=1.5] coordinates {(-42.76, 23.93) (-27.94, 44.51) (-12.58, 65.39) (34.15, -93.29) (11.05, -90.69) (-12.40, -88.02) (-95.38, -105.64) (-74.09, -97.45) (-54.48, -91.10) (92.00, -94.41) (108.00, -78.98) (-71.77, 15.38) (-87.71, 28.73) (-37.91, -59.50) (116.18, -45.40) (-48.90, -15.13)};
      \addplot [myblue,only marks,mark=x,mark size=3.5,line width=1.5] coordinates {(-94.88, -107.34) (-89.18, 29.73) (-73.01, 16.65) (-72.73, -99.19) (-55.71, -93.94) (-48.13, -14.97) (-43.13, 23.12) (-37.57, -59.81) (-29.01, 43.83) (-13.45, -87.06) (-13.68, 65.37) (8.69, -90.57) (34.49, -94.90) (95.11, -95.13) (110.95, -79.55) (115.76, -44.41)};
      \legend{,,opt(16),CNN(16)}
      \end{axis}
    \end{tikzpicture}
  }
  \hfil
  \subfloat[][]{
    \begin{tikzpicture}
      \begin{axis}[title={opt. (0.005, 0) | CNN (0.008, 0)}, title style={at={(0.50,0.8)}, font=\scriptsize}, legend style={cells={anchor=west}, font=\scriptsize}, legend columns=4, legend pos=south west, height=\linecircleimgwidth\textwidth, width=\linecircleimgwidth\textwidth, ticks=none, scale only axis, xmin=-160, xmax=160, ymin=-160, ymax=160,enlargelimits=false,axis on top,]
      \addplot graphics [xmin=-160,xmax=160,ymin=-160,ymax=160] {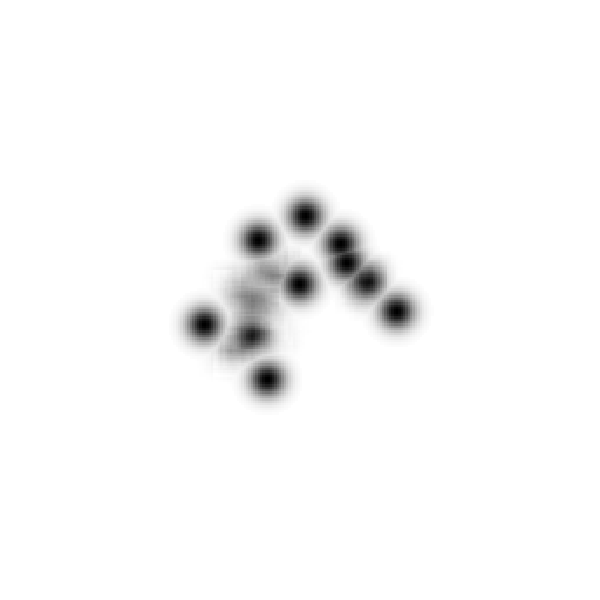};
      \addplot [red,only marks,mark size=1.7] coordinates {(-17.39, -42.40) (51.67, -6.17) (-22.09, 31.82) (21.80, 29.83) (-50.97, -13.20) (24.90, 19.74) (2.96, 44.79) (-0.20, 8.45)};
      \addplot [mygreen,only marks,mark=o,mark size=2.5,line width=1.5] coordinates {(-36.31, 9.80) (-36.74, -26.01) (35.28, 10.38)};
      \addplot [myblue,only marks,mark=x,mark size=3.5,line width=1.5] coordinates {(-25.82, -19.16) (-25.55, -1.26) (35.74, 9.57)};
      \legend{,task,opt(3),CNN(3)}
      \end{axis}
    \end{tikzpicture}
  }
  \caption{\label{fig:gen_examples}Generalization examples for 8 and 12 agent teams. See Figs. \ref{fig:line_test}, \ref{fig:circle_test} for complete description. Images are 256x256 covering an area 320 meters on a side.}
\end{figure}

Indeed, we find this to be the case. We applied our ConvAE network trained exclusively on 2-6 task agent teams to a test dataset with much larger teams of 8 and 12 task agents with 3,000 and 1,500 samples, respectively. Fig. \ref{fig:dataset_stats_812} shows that our model generalizes well to these out-of-distribution test cases. Representative results are show in Fig. \ref{fig:gen_examples}. We note that this is not a trivial result. To achieve good performance with our method on large teams one need only train ConvAE on a comprehensive dataset of smaller configurations and fine tune on a much smaller set of larger team samples.

\subsection{Scalability}
\label{sec:org9641884}
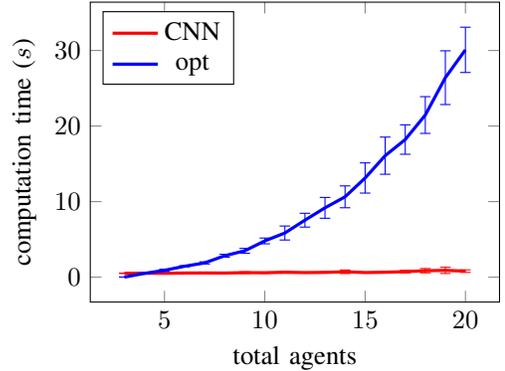
\begin{figure}
  \centering
  \begin{tikzpicture}
    \begin{axis}[
      height = 0.22\textwidth,
      width = 0.30\textwidth,
      scale only axis,
      xlabel = {total agents},
      ylabel = {computation time ($s$)},
      legend pos=north west,
    ]
    \addplot+ [color=red,
               very thick,
               solid,
               mark=.,
               error bars/.cd,
                   y dir=both,y explicit,
    ] table [x=x,y=y,y error=error] {
        x   y       error
        3   0.5197  0.0256
        4   0.5314  0.0244
        5   0.5164  0.0321
        6   0.5555  0.0308
        7   0.5629  0.0233
        8   0.5517  0.0346
        9   0.6064  0.1038
        10  0.5935  0.0721
        11  0.6584  0.1219
        12  0.6122  0.0389
        13  0.6383  0.0626
        14  0.7220  0.2359
        15  0.6187  0.0335
        16  0.6576  0.0869
        17  0.7360  0.1666
        18  0.8617  0.2867
        19  0.9079  0.4119
        20  0.8050  0.1656
    };
    \addplot+ [color=blue,
               very thick,
               solid,
               mark=.,
               error bars/.cd,
                   y dir=both,y explicit,
    ] table [x=x,y=y,y error=error] {
        x   y       error
        3   0.0181  0.0010
        4   0.5052  0.0758
        5   0.9129  0.1488
        6   1.4478  0.0804
        7   1.8924  0.1448
        8   2.8414  0.1966
        9   3.4963  0.3229
        10  4.8038  0.3818
        11  5.8490  0.9252
        12  7.5484  0.9142
        13  9.1807  1.3787
        14  10.638  1.4460
        15  13.131  1.9976
        16  16.084  2.4575
        17  18.219  1.9471
        18  21.463  2.4311
        19  26.405  3.5568
        20  30.089  2.9983
    };
    \legend{CNN, opt}
    \end{axis}
  \end{tikzpicture}
  \caption{\label{fig:computation_time}A comparison of the computation time required to find relay node positions using our CNN approach compared with the optimization-based approach in Eq. \eqref{eq:connectivity_sdp} (opt).}
\end{figure}

A main motivation for our work is scalability. While Eq. \eqref{eq:connectivity_sdp} offers an elegant solution to the connectivity problem it becomes prohibitively slow as the size of the team grows and thus is unsuited for real-time deployment. On the other hand, our approach built on CNN inference is exceptionally scalable. Fig. \ref{fig:computation_time} shows a comparison of computation time between the optimization and CNN. For the CNN we measure the time it takes to convert a task team configuration to a kernelized image (like Fig. \ref{fig:dataset_task_image}), perform inference, and extract the resulting communication team configuration. For the optimization we measure the time it takes to either converge to a local optimum or reach 20 iterations. Note that it often takes more than 20 iterations for the optimization to converge; however, to account for often brittle convergence criteria we impose an aggressive upper limit on the maximum number of optimization iterations for the purpose of comparison. The advantage of our learning approach is clear: while the optimization quickly climbs to 10s of seconds the CNN increases from 520ms to only 800ms, with the increase in time coming entirely from unoptimized post processing steps. For a team of 20 agents the CNN is nearly 40 times faster!

\subsection{Dynamic Scenario}
\label{sec:orge379529}
While we have shown the performance of our system across a variety of static test cases, our goal remains to provide \emph{mobile wireless infrastructure on demand} to teams of task agents collaborating to accomplish an objective. To that end, we have also deployed our system in a dynamic scenario in a Unity-based robotics simulator. To adapt our system for online use, we wrap our CNN in a high level controller that takes in the current state of the task agents, marks their positions in an image with a gaussian kernel, and passes it to the CNN for inference. From the image produced by the CNN, the high level controller extracts the target network team configuration using the procedure outlined in Section \ref{sec:org85ae6b9} and sends waypoint commands to the communication agents accordingly.

Our approach is inherently centralized and relies on aggregating state information about each agent in the team at a planning node and disseminating control commands back to the communication agents. This network overhead amounts to transmitting a few floats at each planning step, which need not be run at a high rate. Thus, our method could operate using the network that it provides as a backbone or utilizing a low-bandwidth, long-range control channel. Note also that once deployed our CNN-based planner requires little computation and could easily be run on-board one of the communication agents, alleviating the need for a fixed ground station.

In the dynamic scenario, five task agents patrol a large urban area covering an space approximately 500 meters on a side. These patrolling agents require communication in order to secure their perimeter and thus relay agents are deployed to form a connected network. For this test, the transmit power of the agents was increased to \(21\) dBm (\(d_c \approx 200\) meters) in order to scale their communication range relative to the size of the space. While the CNN was trained on configurations with an underlying transmit power of \(0\) dBm, it can be used directly in this scenario without retraining simply by appropriately scaling the input and output configurations of the CNN. A snapshot of the paths taken by the task and communication agents in this scenario can be seen in Fig. \ref{fig:dynamic_scenario}. During the patrol our CNN-based controller produced solutions to the connectivity problem at a rate of 2 Hz, keeping the patrolling task agents connected the entire time. Contrast that with the optimization approach that takes on the order of \textasciitilde{}7 seconds to come up with a single connected configuration (see Fig. \ref{fig:computation_time}).

One might wonder about the continuity of the target communication team configurations produced by the CNN between iterations. While each image is processed independently, the rate at which the control loop is closed means that the CNN processes relatively small changes in the input image at each time step. Since CNNs are robust to perturbations the target communication team configurations do not change dramatically between iterations, as can be seen in Fig. \ref{fig:dynamic_scenario}. We do note that over time significant changes in the task agent configuration can result in the network topology changing. However, this challenge is faced by every connectivity based method and can be mitigated in practice by planning ahead.

\newcommand{\patrolfigwidth}{0.31}
\newcommand{\patrolmarkersize}{2.5pt}
\begin{figure*}
  \centering
  \includegraphics[trim=200 100 100 65,clip,width=0.37\textwidth]{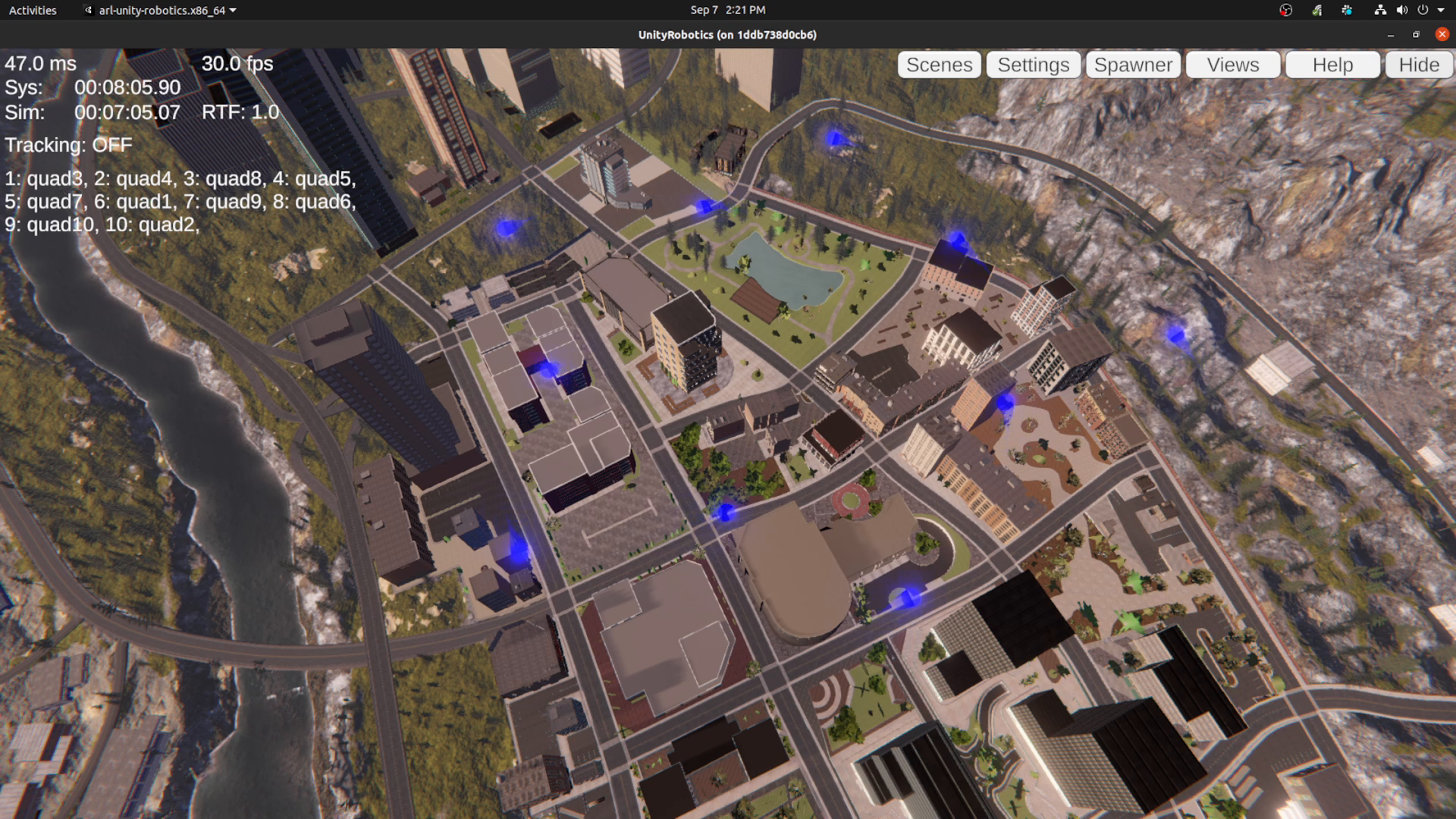}
  \quad
  \begin{tikzpicture}
    \begin{axis}[
      width = \patrolfigwidth*\textwidth,
      axis equal
    ]
    \addplot [color=red,thick,dashed] table [x=q1x,y=q1y,col sep=comma,] {figures/mid_patrol_init.csv};
    \addplot [color=red,thick,dashed] table [x=q2x,y=q2y,col sep=comma,] {figures/mid_patrol_init.csv};
    \addplot [color=red,thick,dashed] table [x=q3x,y=q3y,col sep=comma,] {figures/mid_patrol_init.csv};
    \addplot [color=red,thick,dashed] table [x=q4x,y=q4y,col sep=comma,] {figures/mid_patrol_init.csv};
    \addplot [color=red,thick,dashed] table [x=q5x,y=q5y,col sep=comma,] {figures/mid_patrol_init.csv};
    \addplot [color=blue,thick,dashed] table [x=q6x,y=q6y,col sep=comma,] {figures/mid_patrol_init.csv};
    \addplot [color=blue,thick,dashed] table [x=q7x,y=q7y,col sep=comma,] {figures/mid_patrol_init.csv};
    \addplot [color=blue,thick,dashed] table [x=q8x,y=q8y,col sep=comma,] {figures/mid_patrol_init.csv};
    \addplot [color=blue,thick,dashed] table [x=q9x,y=q9y,col sep=comma,] {figures/mid_patrol_init.csv};
    \addplot [color=blue,thick,dashed] table [x=q10x,y=q10y,col sep=comma,] {figures/mid_patrol_init.csv};
    \addplot [red,only marks,mark=*,mark size=\patrolmarkersize] coordinates {(-184.28,-185.31) (107.96,-174.38) (243.63,16.67) (31.10,208.92) (-219.92,115.40)};
    \addplot [blue,only marks,mark=x,mark size=4pt,line width=1.5] coordinates {(-30.17,-154.44) (-150.50,-35.34) (-72.84,115.49) (97.13,-44.50) (115.78,90.75)};
    \end{axis}
  \end{tikzpicture}
  \begin{tikzpicture}
    \begin{axis}[
      width = \patrolfigwidth*\textwidth,
      axis equal,
      yticklabels={,,}
    ]
    \addplot [color=red,thick,dashed] table [x=q1x,y=q1y,col sep=comma,] {figures/mid_patrol_convae.csv};
    \addplot [color=red,thick,dashed] table [x=q2x,y=q2y,col sep=comma,] {figures/mid_patrol_convae.csv};
    \addplot [color=red,thick,dashed] table [x=q3x,y=q3y,col sep=comma,] {figures/mid_patrol_convae.csv};
    \addplot [color=red,thick,dashed] table [x=q4x,y=q4y,col sep=comma,] {figures/mid_patrol_convae.csv};
    \addplot [color=red,thick,dashed] table [x=q5x,y=q5y,col sep=comma,] {figures/mid_patrol_convae.csv};
    \addplot [color=blue,thick,dashed] table [x=q6x,y=q6y,col sep=comma,] {figures/mid_patrol_convae.csv};
    \addplot [color=blue,thick,dashed] table [x=q7x,y=q7y,col sep=comma,] {figures/mid_patrol_convae.csv};
    \addplot [color=blue,thick,dashed] table [x=q8x,y=q8y,col sep=comma,] {figures/mid_patrol_convae.csv};
    \addplot [color=blue,thick,dashed] table [x=q9x,y=q9y,col sep=comma,] {figures/mid_patrol_convae.csv};
    \addplot [color=blue,thick,dashed] table [x=q10x,y=q10y,col sep=comma,] {figures/mid_patrol_convae.csv};
    \addplot [red,only marks,mark=*,mark size=\patrolmarkersize] coordinates {(88.44, 197.85) (247.77, -47.91) (-195.78, 167.68) (-198.10, -107.15) (33.98, -174.64)};
    \addplot [blue,only marks,mark=x,mark size=4pt,line width=1.5] coordinates {(-184.78, 27.76) (-43.73, 158.68) (141.72, 74.58) (120.88, -83.38) (-68.35, -115.75)};
    \end{axis}
  \end{tikzpicture}
  \caption{\label{fig:dynamic_scenario}A screenshot of the Unity-based robotics simulator showing the location of robots as blue blobs (left) and two snapshots of the dynamic patrol showing the positions of the task agents as red dots and the communication agents following our CNN-based planner as blue x's at the start (center) and middle (right) of the patrol. Dimensions are in meters.}
\end{figure*}
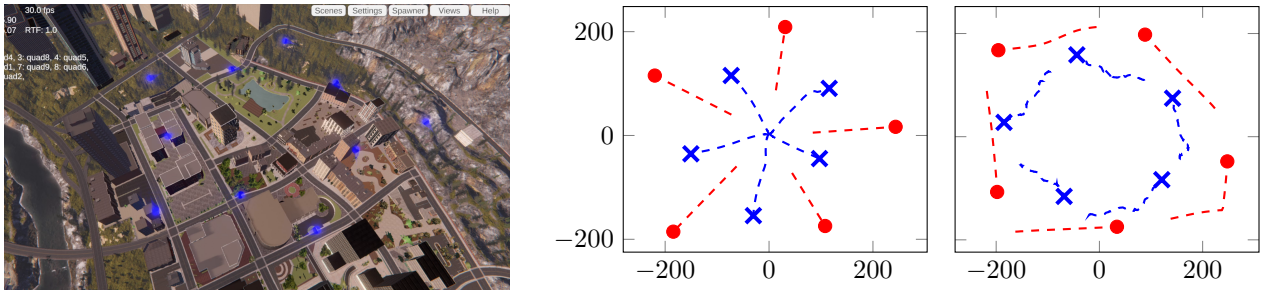

\section{Conclusion}
\label{sec:orga6e01a5}
In this letter we have proposed a data-driven approach to maximizing algebraic connectivity by employing a convolutional autoencoder that learns how to provide \emph{mobile wireless infrastructure on demand} in robot reams requiring communication. While optimization-based methods become slow as the number of agents increases, our CNN-based method scales exceptionally well, running over an order of magnitude faster for large teams. Our system achieves all this while consuming nearly the same amount of transmit power and using an almost equal number of communication agents on average. A natural question is if there is a better network architecture for this specific application that can yield better performance, especially in dynamic scenarios. This is an avenue for future research.

\bibliographystyle{unsrt}
\bibliography{root}
\end{document}